\documentclass[11pt,a4paper]{article}
\usepackage[hyperref]{acl2020}
\usepackage{times}
\usepackage{multirow}
\usepackage{graphicx}  
\usepackage{caption}
\usepackage{booktabs}
\usepackage{enumitem}
\usepackage{subcaption}
\usepackage{latexsym}
\usepackage{soul}
\usepackage{url}

\usepackage{microtype}

\title{Clue: Cross-modal Coherence Modeling for Caption Generation}

\aclfinalcopy
\author{Malihe Alikhani \\
Rutgers University \\\texttt{malihe.alikhani@rutgers.edu} \And
Piyush Sharma \\
Google Research \\\texttt{piyushsharma@google.com}\\\AND
Shengjie Li \\ 
Rutgers University \\\texttt{sl1560@rutgers.edu}\\ \And
Radu Soricut \\
Google Research \\\texttt{rsoricut@google.com}\\\And
Matthew Stone\\
Rutgers University \\\texttt{mdstone@rutgers.edu}}

\date{}

\begin{document}
\maketitle
\begin{abstract}
We use coherence relations inspired by computational models of discourse to study the information needs and goals of image captioning. Using an annotation protocol specifically devised for capturing image--caption coherence relations, we annotate 10,000 instances from publicly-available image--caption pairs.
We introduce a new task for learning inferences in imagery and text, coherence relation prediction, and show that these coherence annotations can be exploited to learn relation classifiers as an intermediary step, and also train coherence-aware, controllable image captioning models.
The results show a dramatic improvement in the consistency and quality of the generated captions with respect to information needs specified via coherence relations.
\end{abstract}

\section{Introduction}
\label{sec:intro}

The task of image captioning is seemingly straightforward to define: use natural language to generate a description that captures the salient content of an image.
Initial datasets, such as MSCOCO \cite{lin2014microsoft} and Flickr \cite{flickr30k}, approached this task directly, by asking crowd workers to describe images in text.
Unfortunately, such dedicated annotation efforts cannot yield enough data for training robust generation models; the resulting generated captions are plagued by content hallucinations \cite{rohrbach2018object,sharma2018conceptual} that effectively preclude them for being used in real-world applications.

In introducing the Conceptual Captions dataset, \newcite{sharma2018conceptual} show that this dataset is large enough, at 3.3M examples, to significantly alleviate content hallucination.
However, because the technique for creating such a large-scale resource relies on harvesting existing data from the web, it no longer guarantees consistent image--text relations.
For example, along with descriptive captions (e.g.,``this is a person in a suit''), the dataset also includes texts that provide contextual background (e.g., ``this is the new general manger of the team'') and subjective evaluations (e.g., ``this is stylish'').
As a result, current captioning models trained on Conceptual Captions avoid content hallucination but also introduce different, more subtle and harder-to-detect issues related to possible context hallucinations (i.e., is this actually the new general manager?) or subjective-judgement hallucinations (i.e., whose judgment is this anyway?).

\begin{figure}
    \centering
    \includegraphics[width=0.9\columnwidth]{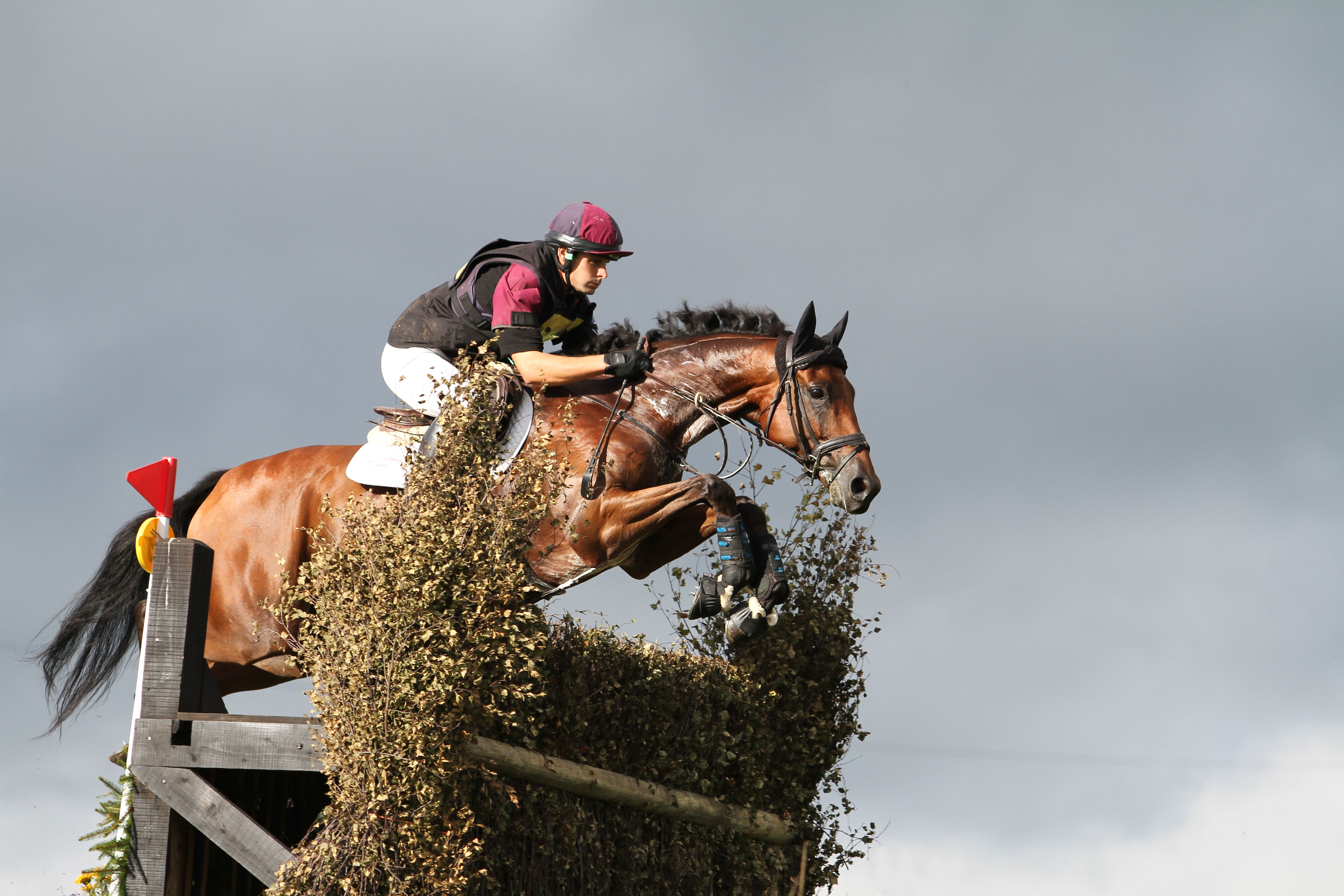}
\caption{Output of a coherence-aware model for various coherence relations.
Content that establishes the intended relation is underlined. (Photo credit: Blue Destiny / Alamy Stock Photo)\\
\textbf{Visible:} horse and rider jumping a fence.\\
\textbf{Meta:} horse and rider jumping a fence \ul{during a race}.\\
\textbf{Subjective:} \ul{the most beautiful} horse in the world.\\
\textbf{Story:} horse \ul{competes} in the event.}
\label{fig:example-intro}
\end{figure}

In this paper, we propose to tackle this issue of large-scale image-caption consistency using a coherence-aware approach inspired by the framework of discourse coherence theory \cite{hobbs1978discourse,phillips1977calculus}.
This framework characterizes the inferences that give discourse units a coherent joint interpretation using a constrained inventory of coherence relations.  In multimodal presentations, discourse units can be images as well as text, so we appeal to new image--text coherence relations that capture the structural, logical, and purposeful relationships between the contributions of the visual modality and the contributions of the textual modality.
For instance, a \textit{Visible} relation characterizes grounding texts that serve to make key aspects of the image content common ground (perhaps to a visually-impaired reader), analogous to \textit{Restatement} relations between one text unit and another; \textit{Visible} relations are key to traditional descriptive captions such as ``this is a person in a suit.''  Meanwhile, a \textit{Story} relation characterizes texts that develop the circumstances depicted in the image in pursuit of free-standing communicative goals, analogous to \textit{Occasion} or \textit{Narration} relations in text; \textit{Story} relations can go far beyond image content (``I hiked this mountain as we found it on a list for good hikes for kids'') and so pinpoint one kind of risk for context hallucinations.  The key contribution of our work is to show that image--text coherence can be systematized, recognized, and used to control image captioning models.

To support our argument, we create a coherence-relation annotation protocol for image-caption pairs, which we use to annotate 10,000 image-caption pairs over images coming from the Conceptual Captions \cite{sharma2018conceptual} and Open Images \cite{kuznetsova2020open} datasets.
We  release\footnote{\url{https://github.com/malihealikhani/Cross-modal_Coherence_Modeling}} this dataset, named Clue, to facilitate follow-up research.
By annotating these coherence relations in the context of image captioning, we open up the possibility of analyzing patterns of information in image--text presentations at web scale.

In addition, we show that we can exploit these coherence-relation annotations by training models to automatically induce them, as well as by building models for coherence-aware image captioning.
Because they are driven by input coherence relations, these captioning models can be used to generate captions that are better suited to meet specific information needs and goals.

\section{Prior Work}
\label{sec:related-work}

There are diverse ways to characterize the communicative functions of text and images in multi-modal documents \cite{marsh2003taxonomy}, any of which can provide the basis for computational work.  Some studies emphasize the distinctive cognitive effects of imagery in directing attention; engaging perceptual, spatial and embodied reasoning; or eliciting emotion \cite{kruk-etal-2019-integrating,Shuster_2019_CVPR}.  Some look at contrasts across style and genre \cite{guo2019mscap}.  Others look holistically at the content of text and imagery as complementary or redundant \cite{otto2019understanding,vempala2019categorizing}.  Unlike our approach, none of these methodologies attempt to characterize information-level inferences between images and text, so none is suitable for building generation models that control the information that text provides.

While coherence theory has been applied to a range of multimodal communication, including comics \cite{mccloud1993understanding}, gesture \cite{lascarides2009formal}, film \cite{cumming2017conventions}, and demonstrations and other real-world events \cite{hunter:etal:2018,stojnic:pp13}, applying coherence theory specifically to text--image presentations is less well explored.  The closest work to ours is \newcite{alikhani2019cite}, who explore coherence relations between images and text in a multimodal recipe dataset.  Their relations are specialized to  instructional discourse and they do not build machine learning models combining imagery and text. We consider more general coherence relations and a broader range of machine learning methods. 

We use our relations and introduce a coherence-aware caption generation model that improves the rate of good \textit{Visible} captions by around 30\%. This is a considerable improvement over the recent models that have tried to achieve more control over neural language generation using an enhanced beam search \cite{anderson-etal-2017-guided}, a memory network with multiple context information \cite{Park_2017_CVPR}, forced attentions \cite{sadler-etal-2019-neural} and modeling and learning compositional semantics using fine-grained annotations of entities in MSCOCO \cite{cornia2019show}.

\begin{figure*}[ht!]
\centering
\begin{subfigure}[t]{0.23\linewidth}
\centering
\caption*{Visible, Meta}
\includegraphics[width=0.90\linewidth, height= 2.7cm]{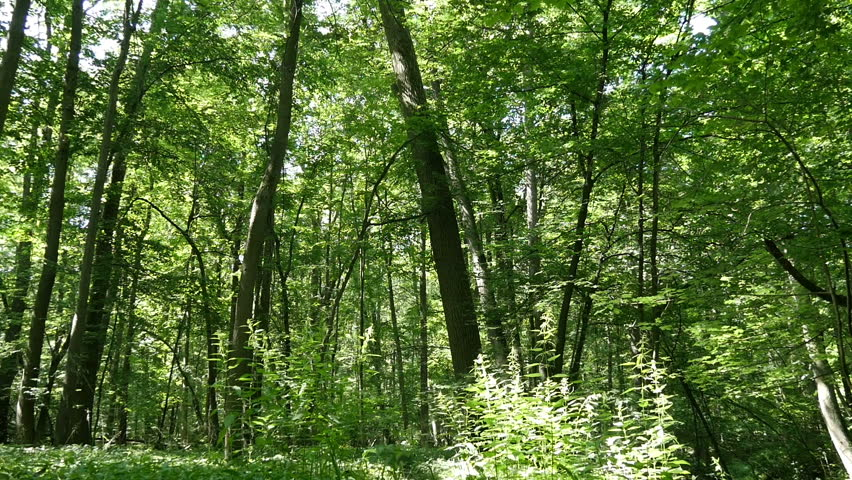}
\caption{\textsc{Caption:} forest on a sunny day}
\end{subfigure}
\hspace{5pt}
\begin{subfigure}[t]{0.23\linewidth}
\centering
\caption*{Visible, Action, Subjective}
\includegraphics[width=0.99\linewidth, height= 2.5cm]{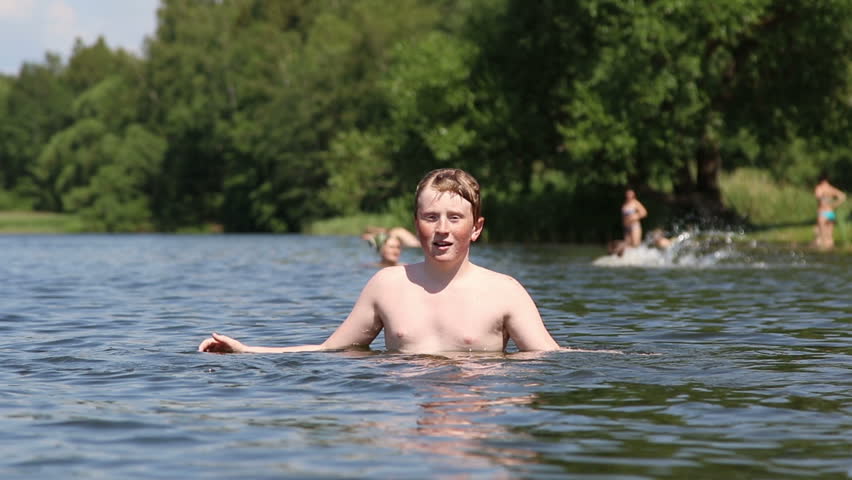}
\caption{\textsc{Caption:} young happy boy swimming in the lake.}
\end{subfigure}
\hspace{5pt}
\begin{subfigure}[t]{0.23\linewidth}
\centering
\caption*{Meta, Action, Story}
\includegraphics[width=\linewidth, height= 2.5cm]{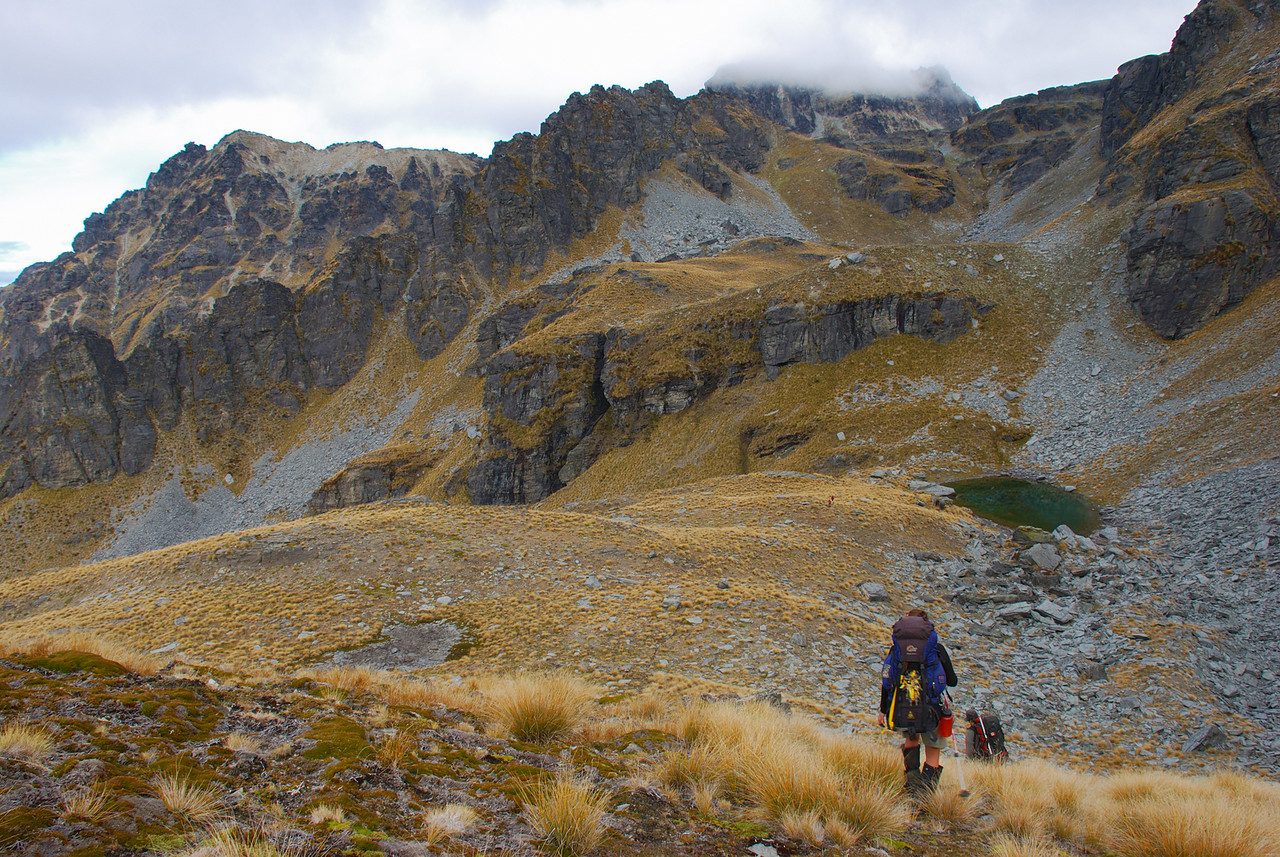}
\caption{\textsc{Caption:}  approaching our campsite, at 1550m of elevation on the slopes.}
\end{subfigure}
\hspace{5 pt}
\begin{subfigure}[t]{0.23\linewidth}
\centering
\caption*{Irrelevant}
\includegraphics[width=\linewidth, height= 2.5cm ]{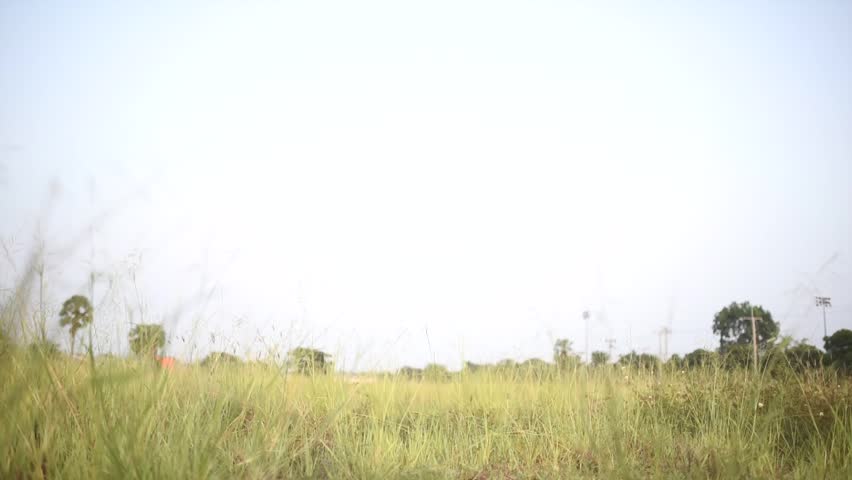}
\caption{\textsc{Caption:}  young girl walking on the dry grass field under daylight.}
\end{subfigure}
\caption{ We use a constrained set of coherence relations to summarize the structural, logical and purposeful relationships between the contributions of text and the contributions of images. Multiple coherence relations can be found simultaneously.  (Image--caption pairs are chosen from the Conceptual Caption dataset; photo credits: Dmytro Zinkevych; Shutterstock user yauhenka; Danilo Hegg; Andre Seale)}
\label{fig:relations}
\end{figure*}

\section{Coherence in Images and Captions}
\label{sec:coherence}

The first step toward our goals is to characterize image--text coherence and annotate a sizable corpus of image--text pairs with coherence relations.

We use an overlapping set of high-level relations, inspired both by theoretical work linking discourse coherence to discourse structure and discourse goals \cite{roberts2012information,webber1999discourse}, and by previous successful discourse annotation campaigns \cite{prasad2008penn}.  Crucially, following previous work on text \cite{rohde-etal-2018-discourse} and multimodal discourse \cite{alikhani2019cite}, we assume that several of these relations can hold concurrently.  The relations are:
\begin{itemize}[noitemsep,nolistsep] 
\item
\textit{Visible}, where text presents information that is intended to recognizably characterize what is depicted in the image, analogous to \textit{Restatement} relations in text \cite{prasad2008penn}.
\item
\textit{Subjective}, where the text describes the speaker's reaction to, or evaluation of, what is depicted in the image, analogous to \textit{Evaluation} relations in text \cite{hobbs1985coherence}; 
\item
\textit{Action}, where the text describes an extended, dynamic process of which the moment captured in the image is a representative snapshot, analogous to \textit{Elaboration} relations in text \cite{prasad2008penn};
\item
\textit{Story}, where the text is understood as providing a free-standing description of the circumstances depicted in the image, analogous to the \textit{Occasion} relation of \newcite{hobbs1985coherence} but including instructional, explanatory and other background relations; and
\item 
\textit{Meta}, where the text allows the reader to draw inferences not just about the scene depicted in the image but about the production and presentation of the image itself, analogous to \textit{Meta-talk} relations in text \cite{doi:10.1111/j.1475-682X.1980.tb00021.x}.
\end{itemize}

Figures~\ref{fig:relations}(a), (b) and (c) show examples of image--caption pairs and the associated coherence relations. We can see that image--caption pairs often have multiple relations.
For completeness, we also present in Figure~\ref{fig:relations}(d) an example of an image--caption pair that does not fall into any of the above categories (and it is therefore labeled \textit{Irrelevant}).

\subsection{Data Collection}

Clue includes a total of 10,000 annotated image--caption pairs.
A first subset of 5,000 image--caption pairs was randomly selected from the training split of the Conceptual Captions dataset \cite{sharma2018conceptual}, as a representative sample of human-authored image captions.
The Conceptual Captions dataset is a collection of web-harvested images paired with their associated ALT-TEXT, created by human authors under various non-public guidelines (regarding style, objectivity, etc.) for over 111,000 web pages including news articles, advertisements, educational posts, blogs, etc.

A second subset of 5,000 image--caption pairs, to be used as a representative sample of machine-authored captions, is obtained from the outputs of 5 of the top models that participated in the image-captioning challenge for the Conceptual Caption Workshop at the 2019 Conference on Computer Vision and Pattern Recognition (CVPR).
These machine-authored captions are over a set of 1,000 images from the Open Images Dataset~\cite{kuznetsova2020open}, and are publicly available.\footnote{http://www.conceptualcaptions.com/winners-and-data}

\paragraph{Protocol}
Although specific inferences have been shown to be realizable by crowd workers \cite{alikhani2019cite}, the results of our pilot studies for annotating these more general relations with the help of crowd workers were not satisfactory.
We have found that expert raters' decisions, however, have high agreement on our discourse categories. 
%
The study has been approved by Rutgers's IRB; the annotators, two undergraduate linguistics students, were paid a rate of \$20/h.

In our annotation protocol, we ask the annotators to label the main relations described in Section~\ref{sec:coherence}, as well as certain fine-grained sub-relations.
The following briefly summarizes our guidelines;
our GitHub repository includes an exact copy of what the annotators used.

Annotations of \textit{Visible} are given for captions that present information intended to recognizably characterize what is depicted in the image, while annotations of \textit{Meta}
indicate not only information about the scene depicted but also about the production and presentation of the image itself.
The \textit{Meta} labels have additional fine-grained labels such as \textit{When}, \textit{How}, and  \textit{Where}.
A few details regarding these fine-grained labels are worth mentioning:
location mentions such as ``in the city'' are labeled as \textit{Meta–-Where}, but generic states, e.g., ``in the snow,'' are merely annotated as \textit{Visible}.
Captions considering the view or the photo angles, or a photo’s composition, i.e. ``portrait'' or ``close-up'', are annotated as \textit{Meta–-How}.

Annotations of \textit{Subjective} are primarily given for captions that included phrases with no objective truth value, i.e. phrases using predicates of personal taste.
For example, captions including noun phrases like ``pretty garden'' are annotated as \textit{Subjective}: whether the garden is pretty or not cannot be determined except by appeal to the opinions of an implicit judge. Note that first-person reports, like ``I want ...'' or ``I need ..." are not annotated as \textit{Subjective} but rather as \textit{Story}, because they describe the speaker's definite state rather than an implicit judgment. 

Captions annotated as \textit{Story} cover a much wider range compared to captions in other categories, including \textit{Meta} and \textit{Subjective}.
These captions range from those that read like instructions, i.e. ``how to ...", to those that present speaker desires, i.e. ``I want ...'' or ``I need ...", to those that give background information not captured in the image, i.e. ``she is an actress and model", and more. 

\begin{figure}[ht]
\centering
\begin{subfigure}[t]{0.45\linewidth}
\centering
\caption*{Other--Text}
\includegraphics[width=\linewidth]{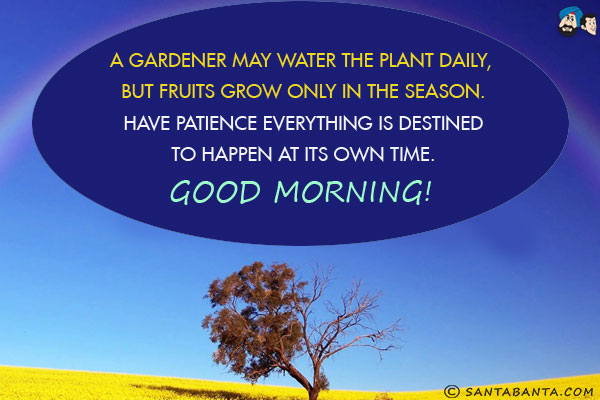}
\caption{\textsc{Caption:} a gardener may water the plant daily but fruits grow only in the season.}
\end{subfigure}
\hspace{1pt}
\begin{subfigure}[t]{0.45\linewidth}
\centering
\caption*{Other--Gibberish}
\includegraphics[width=0.6\linewidth ]{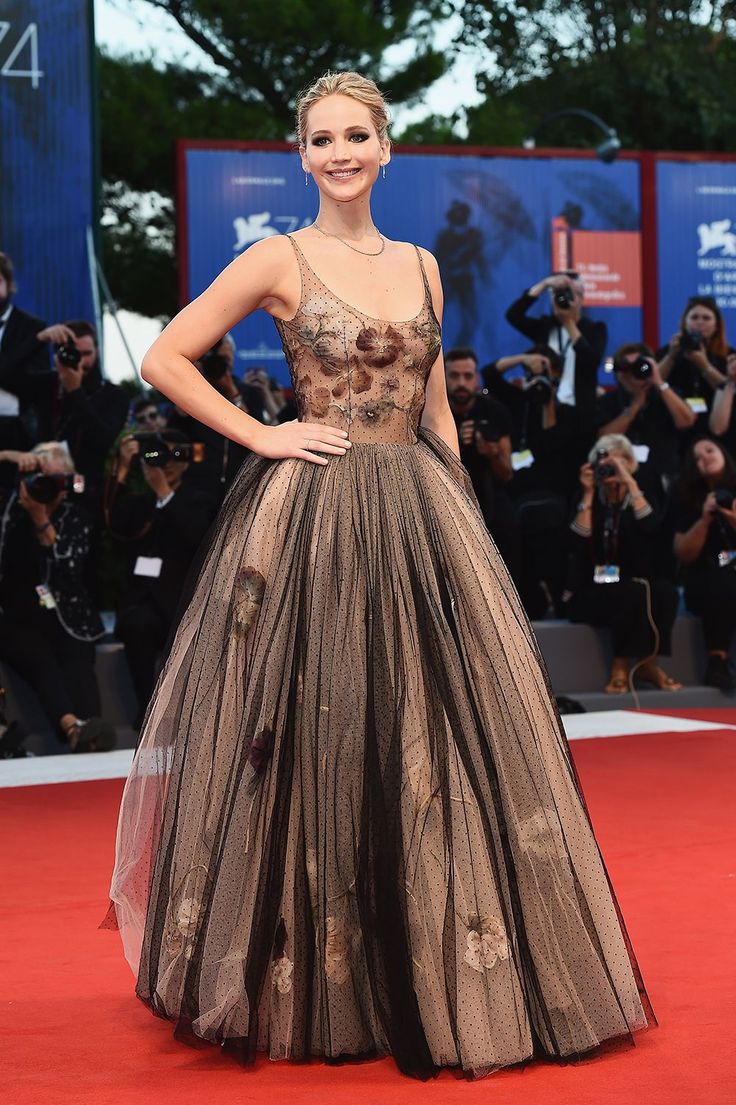}
\caption{\textsc{Caption:} actor in retail at the mother.}
\end{subfigure}
\caption{Examples of image--caption pairs in the \textit{Other}
category. (Photo credit: santabanta.com; Mary Sollosi)}
\label{fig:errors}
\end{figure}

\paragraph{\textit{Other} and \textit{Irrelevant}}
Some of these image--caption pairs contain incomplete captions that are hard to understand. A number of these examples include images that contained text. The text in these cases is relevant to the image and the accompanying captions; in this cases, the coherence relations are marked as \textit{Other--Text} (Figure~\ref{fig:errors}). Some examples of such instances are images containing signs with text, greetings on cards, or text that does not affect the interpretation of the image or caption, such as city names or watermarks.

Other times, the caption text is irrelevant and indicate that the image and caption do not correlate.
Some examples of these instances are captions of ``digital art selected for'' paired with an irrelevant image, and images that clearly do not match the caption, such as an image of a man walking with the caption ``a field of strawberries''.
We have specifically labeled cases where the caption is almost true or almost relevant to the image at hand, such as the caption ``horses in a field'' with an image containing donkeys with ``minor error". Other cases include images that look like powerpoint slides with bullets and text.
Our GitHub repository includes detailed examples and explanations.

\paragraph{Experiment Interface}
We have developed software for annotating coherence relations in image--text presentations that can flexibly and easily accommodate various annotation schema. The annotators used this software for annotating the image--text pairs. They had the option of choosing multiple items and leaving comments. 

\paragraph{Agreement}
To assess the inter-rater agreement, we determine Cohen’s $\kappa$.
For this, we randomly chose 300 image--caption pairs from the Conceptual Caption ground-truth data and assigned them to two annotators.
The resulting $\kappa$ coefficient is $0.81$, which indicates a high agreement on these categorical decisions.

\begin{table*}[ht!]
\centering
\begin{tabular}{@{}cllllll@{}}
\toprule
                     & \multicolumn{1}{c}{Visible} & Subjective & Action  & Story &  Meta  & Irrelevant \\ \midrule
Ground-truth         & 64.97\%                  & 9.77\% & 18.77\% &  29.84\% &  24.59\%  &3.09\%     \\
Model output        & 69.72\%                  & 1.99\% & 11.22\% &  17.19\% &  58.94\%   &16.97\%    \\\bottomrule
Ground-truth + Model & 66.91\%                  & 6.58\%  & 15.68\% & 24.67\% &  38.65\%  &8.77\%     \\ \bottomrule
\end{tabular}
\caption{Distribution of coherence relations over the ground-truth and the model outputs.}
\label{tab:stats}
\end{table*}

\begin{table}[ht!]
\centering
\resizebox{0.95\columnwidth}{!}{%
\begin{tabular}{clll}
\toprule
                     & \multicolumn{1}{c}{When} & How   & Where \\ \hline
\small{Ground-truth}         & 33.74\%                      & 64.40\%   & 28.60\%   \\
\small{Model output}         & 21.75 \%                     & 72.84\%   & 41.03\%   \\\bottomrule
\end{tabular}}
\caption{Distribution of fine-grain relations in the Meta category over the ground-truth and the model outputs.}
\label{tab:when-how-where}
\end{table}

\subsection{Analysis}
In this section we present the overall statistics of the dataset annotations, the limitations of the caption-generation models, and the correlation of the distribution of the coherence relations with genre.

\paragraph{Overall statistics}
The exact statistics over the resulting annotations are presented in Table~\ref{tab:stats} and Table~\ref{tab:when-how-where}.
Overall, \textit{Visible} captions constitute around 65\% and 70\% of captions for the ground-truth labels and the model outputs, respectively.
The rate of \textit{Subjective} and \textit{Story} captions decreases significantly for the model outputs (compared to ground-truth), indicating that the models learn to favor the \textit{Visible} relation at the expense of \textit{Subjective} and \textit{Story}.
However, the rate of \textit{Meta} captions increases by around 25\% in the model outputs,
which points to potential context hallucination effects introduced by these models.
As expected, the rate of \textit{Irrelevant} captions increases to around 17\% in the model-generated captions, compared to 3\% in the ground-truth captions.   
Moreover, it appears that the models have some ability to learn to generate the locations that events take place; however, there is a drop in their ability to generate temporal information (see Table~\ref{tab:when-how-where}). 

In terms of overlap, \textit{Visible} and \textit{Meta} overlap 22.49\% of the time for the ground-truth captions, whereas this rate goes up to 54.55\% in the model outputs.
This ``conflation'' of these two relations is highly problematic, and one of the main motivations for building caption-generation models that have control over the type of discourse relation they create (see Section~\ref{sec:generation}).
Our GitHub page includes additional statistics about overlapping relations.

\begin{figure}[h!]
\centering
\includegraphics[width=0.99\columnwidth]{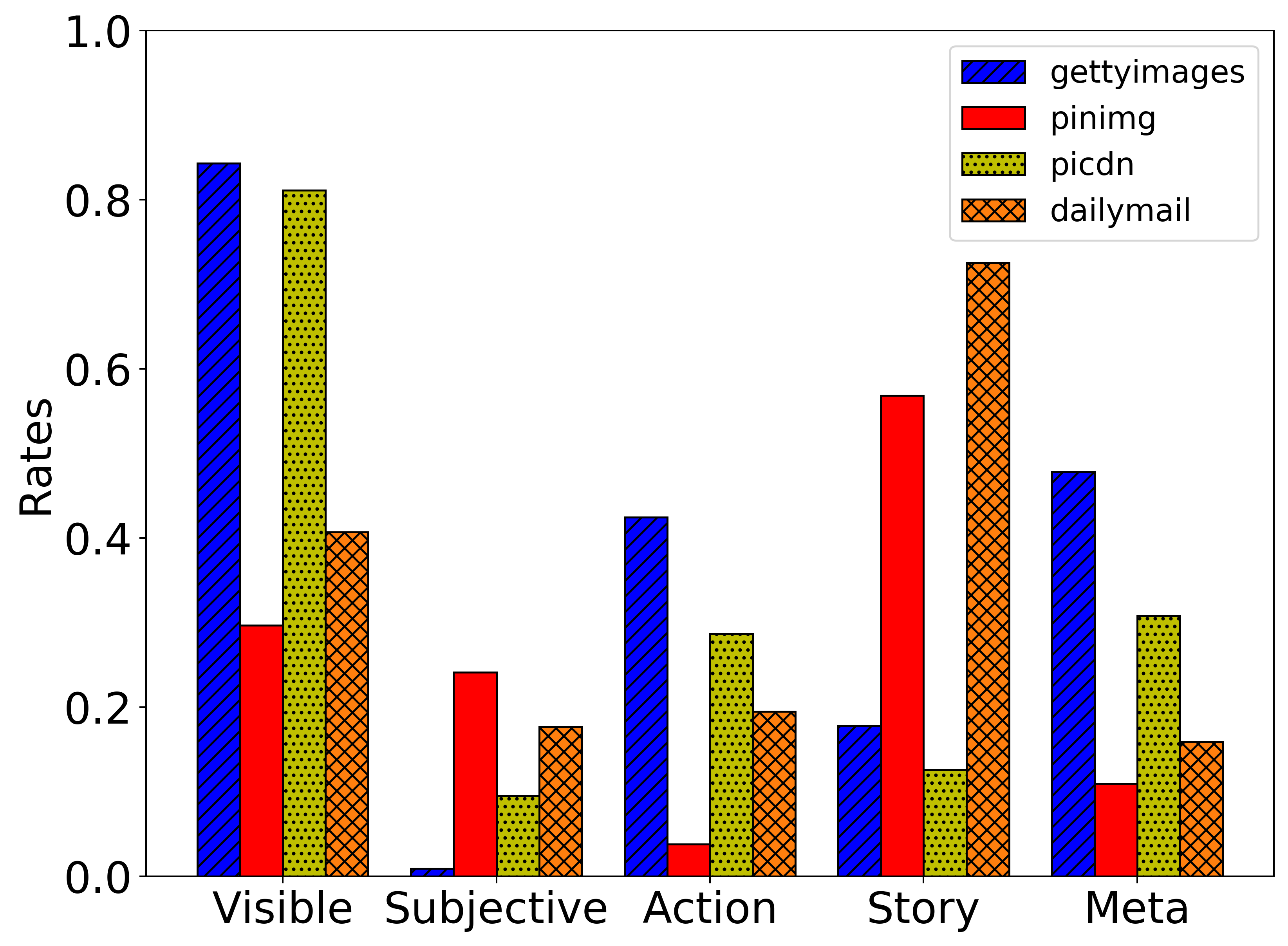}
\caption{Different resources have different kinds image--caption pairs. The graph shows the distribution of labels in the top four domains present in the Conceptual Captions dataset.}
\label{fig:website-dist}
\end{figure}

\paragraph{Coherence relations indicate Genre}
Coherence relations are indicative of the discourse type and its goals, and therefore our annotations correlate with the genre under which the captions have been produced.
That is, image--caption pairs from different publication sources have different distributions of coherence relations.
For instance, pairs from the Getty Images domain mostly come with the \textit{Meta} and \textit{Visible} relations.
In contrast, from the Daily Mail domain are mostly story-like, and include very few captions that describe an action, compared with the Getty Images and picdn domains.
Figure~\ref{fig:website-dist} shows the distribution of the coherence labels for the top four domains from the Conceptual Caption dataset.

\section{Predicting Coherence Relations}
\label{sec:classification}


\begin{table*}[ht!]
\centering
\begin{tabular}{@{}l|llllll|l@{}}
\toprule
                                        & \small{Visible} & \small{Subjective} & \small{Action} & \small{Story} & \small{Meta} & \small{Irrelevant} & \small{Weighted} \\ \midrule
\small{SVM (text-only)}            & 0.83                             & 0.12                                & 0.32                            & 0.21                           & 0.19                          & 0.00                                & 0.48                              \\
\small{GloVe (text-only)}         & 0.80                             & 0.44                                & 0.58                            & 0.57                           & 0.44                          & 0.08                                & 0.63                              \\
\small{BERT (text-only)}          & 0.82                             & 0.35                                & 0.62                            & 0.62                           & 0.44                          & 0.06                                & 0.65                              \\
\small{GloVe + ResNet} & 0.81            & 0.36               & 0.58           & 0.60           & 0.45         & 0.07               & 0.64                              \\
\small{BERT + ResNet}  & 0.83                             & 0.36                                & 0.69                            & 0.62                           & 0.44                          & 0.06                                & 0.67                              \\ \bottomrule
\end{tabular}
\caption{The $F_1$ scores of the multi-class classification methods described in Section~\ref{sec:classification-multi}; 80-20 train-test split; 5-fold cross validation.}
\label{tab:results-classification}
\end{table*}

In this section, we introduce the task of predicting cross-modal coherence relations. We describe a number of preliminary experiments that justify the potential of machine learning models in classifying coherence relations in text and imagery. To this end, we train and test different models on the Clue dataset to automatically predict the coherence labels given an image and its caption.

\subsection{Multi-Label Prediction}
\label{sec:classification-multi}
We first treat the relation prediction problem in its original multi-label setting.
The train--test split for all the models described in this section is 80\%--20\% and the numbers are reported using 5-fold cross validation.

As a baseline, we report the results of a SVM classifier that uses only the text to predict the relationship between image-caption pairs.
We extract bag-of-words features by using N-grams (for N from 1 to 5), and pass them to the SVM classifier as input.
Next, we discuss two multi-modal classifiers for predicting the image--caption coherence relations. 

\paragraph{GloVe + ResNet-50}\label{para:glove}
This model contains a text encoder for textual-feature extraction and an image encoder for image-feature extraction. 
For the image encoder, we use a ResNet-50 \cite{DBLP:conf/cvpr/HeZRS16} pre-trained on ImageNet followed by a Batch-Norm layer, a fully connected layer and a ReLU activation function. The text encoder takes as input word embeddings from the GloVe model \cite{pennington2014glove}, and consists of an LSTM layer, a Batch-Norm layer, a fully connected layer with tanh activation function. 

\paragraph{BERT + ResNet-50}
To test the impact of the text encoder in this setup, we reuse the setup of the previous model with a different textual-feature extractor.
We train and test using an encoder that takes sentence embeddings as input using the $\langle$CLS$\rangle$ representation produced by the BERT-base model~\cite{devlin2018bert}.

\paragraph{Results}
The results of all of our models are presented in Table~\ref{tab:results-classification},
where we present the $F_1$ scores over each of the individual relations, as well as an overall weighted average.
The BERT+ResNet model achieves the highest performance ($|t|> 9.54$, $p<0.01$), with an overall $F_1$ score of $0.67$.
For the interested reader, we present in the GitHub page the top features of the Naive Bayes SVM classifier \cite{wang2012baselines}.

\begin{table*}[ht!]
\centering
\begin{tabular}{l|llllll|l}
\toprule
                                                         & \small{Visible} & \small{Subjective} & \small{Action} & \small{Story} & \small{Meta} & \small{Irrelevant} & \small{Weighted} \\ \midrule
                                                         \small{Ground-truth Distribution}    & 46.65\%  & 7.07\% & 1.31\% &  19.09\%  &  23.42\% & 2.46\% &      \\ \hline

\small{BERT + ResNet}   & 0.64                             & 0.26                                & 0.02                            & 0.52                           & 0.46                          & 0.07                 & 0.52                              \\
\small{BERT + GraphRise}   & 0.59                             & 0.15                                & 0.00                            & 0.42                           & 0.34                          & 0.00                 & 0.45                              \\
\small{USE + GraphRise} & 0.69                             & 0.45                                & 0.00                            & 0.57                           & 0.48                          & 0.00                 & \textbf{0.57}                              \\ \bottomrule
\end{tabular}
\caption{The $F_1$ scores of coherence relation classifiers \textbf{with label mapping}. The aggregated Weighted scores use the numbers in the first row as weights.}
\label{tab:results-classification-label-mapping}
\end{table*}

\subsection{Single-Label Prediction}
To achieve the goal of generating captions with a desired coherence relation to the image, it is important to clearly differentiate between often co-occurring label types (such as \textit{Visible} and \textit{Meta}).
To this end, we introduce a label-mapping strategy for predicting coherence relations, such that each image--caption pair is assigned a single coherence label.
We map the set of human-annotated coherence relations for an image--caption pair to a single label using the following heuristic:

\begin{enumerate}[noitemsep,nolistsep] 
    \item {If the set contains the \textit{Meta} label, then the image--caption pair is assigned the \textit{Meta} label.}
    \item {If the set contains the \textit{Visible} label and does not contain either \textit{Meta} or \textit{Subjective}, then the image--caption pair is set to \textit{Visible}.}
    \item {If none of the above rules are met for this image--caption pair, we randomly sample a label from its set of labels.}
\end{enumerate}

The distribution of labels after this mapping is given in the first row of Table \ref{tab:results-classification-label-mapping}.
As opposed to the ground-truth label distribution in Table~\ref{tab:stats}, these values add up to 100\%.


Using the label mapping described above, we retrain and evaluate the BERT+ResNet classifier presented in Sec.~\ref{sec:classification-multi}.  
In addition, we perform additional experiments in which the caption text in encoded using the pre-trained Universal Sentence Encoder\footnote{tfhub.dev/google/universal-sentence-encoder-large/3}~(USE)~\cite{cer-etal-2018-universal}, which returns a 512-dimensional embedding for the text.
On the image encoding side, we also experiment with the pre-trained Graph-Regularized Image Semantic Embedding model \cite{10.1145/3336191.3371784}, which is trained over ultra-fine--grained image labels over web-sized amounts of data -- roughly 260M examples over roughly 40M labels; this model returns a compact, 64-dimensional representation for the image.
We concatenate the text and image features into a single vector, and feed it to a fully-connected neural network with 3 hidden layers of 256 units each with ReLU activations (for all but the last one), followed by a softmax layer which computes the logits for the 6 target classes.
We divide the 3910 labeled image--text pairs from the ground-truth split of our data into training and test sets, with 3400 and 510 samples, respectively.
We use dropout with probability of 0.5, and tune the model parameters using the Adam optimizer~\cite{kingma2014adam} with a learning rate of $10^{-6}$.

\paragraph{Results}
Table~\ref{tab:results-classification-label-mapping} shows  the results of the single-label prediction experiments, where we present the $F_1$ scores over each of the individual relations, as well as an overall weighted average.
%
%
%
The USE+GraphRise model using the label mapping achieves the highest performance, with an overall $F_1$ score of 0.57. 
Next, we describe how we use this classifier's predictions to annotate the training and validation splits of the Conceptual Caption dataset (3.3 million image--captions pairs), in order to train a controllable caption-generation model.

\begin{table*}[ht!]
\centering
\begin{tabular}{l|l|llll}
\toprule
                               &\begin{tabular}[c]{@{}l@{}}Coherence\\ agnostic\end{tabular} & \begin{tabular}[c]{@{}l@{}}Visible\\ \small{coherence-aware}\end{tabular}  & \begin{tabular}[c]{@{}l@{}}Subjective\\ \small{coherence-aware}\end{tabular} & \begin{tabular}[c]{@{}l@{}}Story\\ \small{coherence-aware}\end{tabular} & \begin{tabular}[c]{@{}l@{}}Meta\\ \small{coherence-aware}\end{tabular} \\ \hline
\multicolumn{1}{l|}{Visible}     & \textbf{52.1\%}    & \textbf{79.9\%}    &  31.7\%   & 25.0\%    & 42.80\%           \\
\multicolumn{1}{l|}{Subjective} & 11.4\%  & 2.6\%  & 24.4\%  & 2.6\%   & 1.9\%        \\
\multicolumn{1}{l|}{Action}     & 10.7\%  & 10.8\% & 6.3\%     & 8.8\%     & 11.4\%        \\
\multicolumn{1}{l|}{Story}     & \textbf{51.3\%} & 16.0\%   & \textbf{45.0}\%     & \textbf{58.8\%}     & 17.34\%        \\
\multicolumn{1}{l|}{Meta}      & 31.2\%  & 32.8\% & 15.1\%     & 17.7\%    & \textbf{46.5\%  }         \\
\multicolumn{1}{l|}{Irrelevant} & 12.2\%  & 12.3\%   & 10.7\%   & 9.9\%    & 21.40\%         \\  \hline \hline 
\multicolumn{1}{l|}{When}  & 9.5\% & 5.6\%    & 4.1\%     & 17.7\%    & 9.6\%             \\
\multicolumn{1}{l|}{How}   & 21.3\% & 21.3\%  & 9.6\%        & 25.0\%         & 30.26\%              \\
\multicolumn{1}{l|}{Where} & 5.3\%  & 8.6\%    & 4.1\%    & 8.8\%    & 16.6\%     \\ \bottomrule
\end{tabular}
\caption{The distribution of coherence relations in image--caption pairs when captions are generated with the discourse--aware model vs the discourse agnostic model (the mode of the distribution in bold).}
\label{tab:discourse-aware}
\end{table*}

\section{Generating Coherent Captions}
\label{sec:generation}
We use the coherence label predictions on the Conceptual Captions dataset (Section~\ref{sec:classification}) to train a coherence-aware caption generation model.

\begin{figure}[bh!]
    \centering
    \includegraphics[width=0.95\columnwidth]{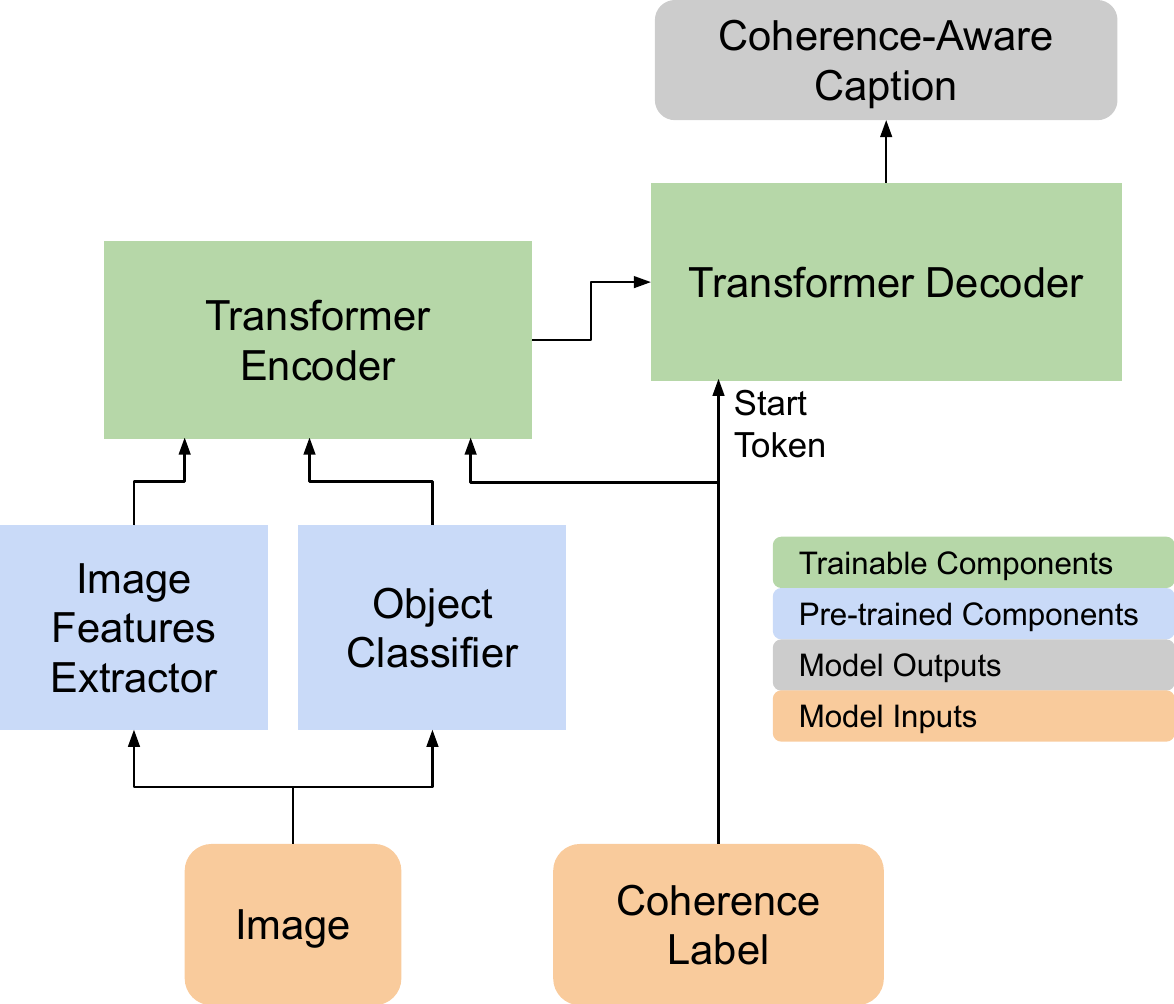}
\caption{Coherence-aware image captioning model}
\label{fig:model-image-captioning}
\end{figure}

\begin{figure*}[th!]
\begin{subfigure}[t]{0.23\linewidth}
\centering
\includegraphics[width=\linewidth]{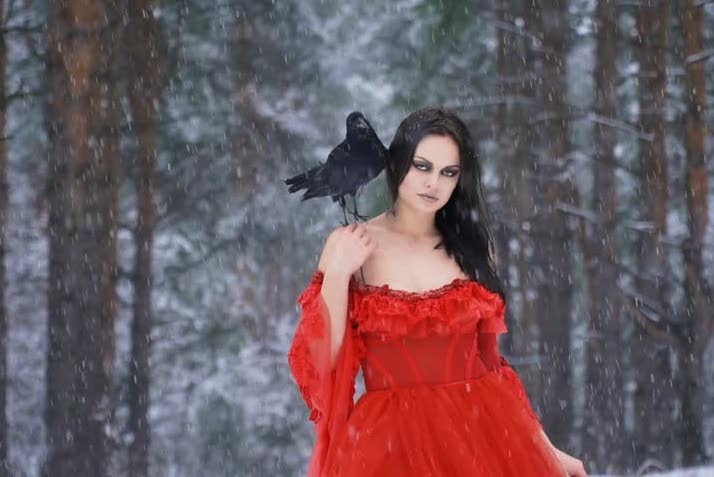}
\caption{coherence-aware \textit{Meta}: A girl in the winter forest.\\ coherence--agnostic: beautiful girl in a red dress.}
\end{subfigure}
\hspace{0.5pt}
\begin{subfigure}[t]{0.24\linewidth}
\centering
\includegraphics[width=\linewidth]{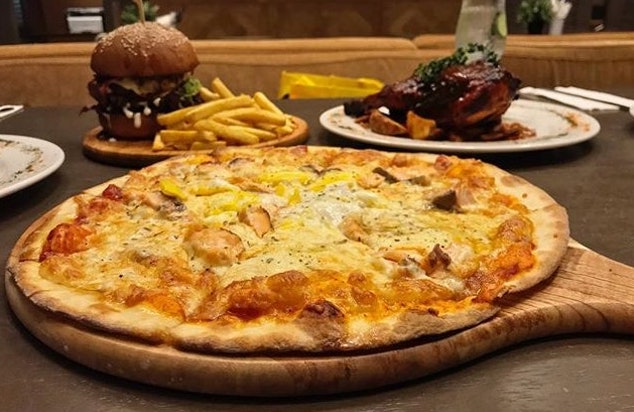}
\caption{coherence-aware \textit{Visible}: the pizza at restaurant is seen.\\ coherence--agnostic: the best pizza in the world.}
\end{subfigure}
\hspace{0.5pt}
\begin{subfigure}[t]{0.23\linewidth}
\centering
\includegraphics[width=\linewidth]{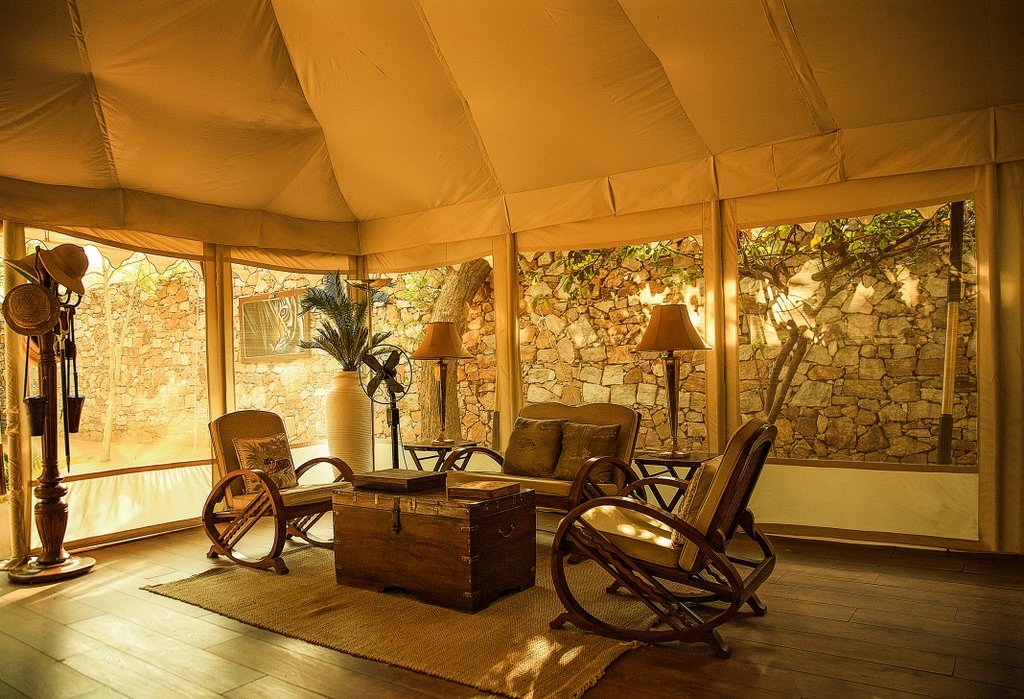}
\caption{coherence-aware \textit{Subjective}: beautiful chairs in a room.\\ coherence--agnostic: the living room of the home. }
\end{subfigure}
\hspace{0.5pt}
\begin{subfigure}[t]{0.24\linewidth}
\centering
\includegraphics[width=\linewidth]{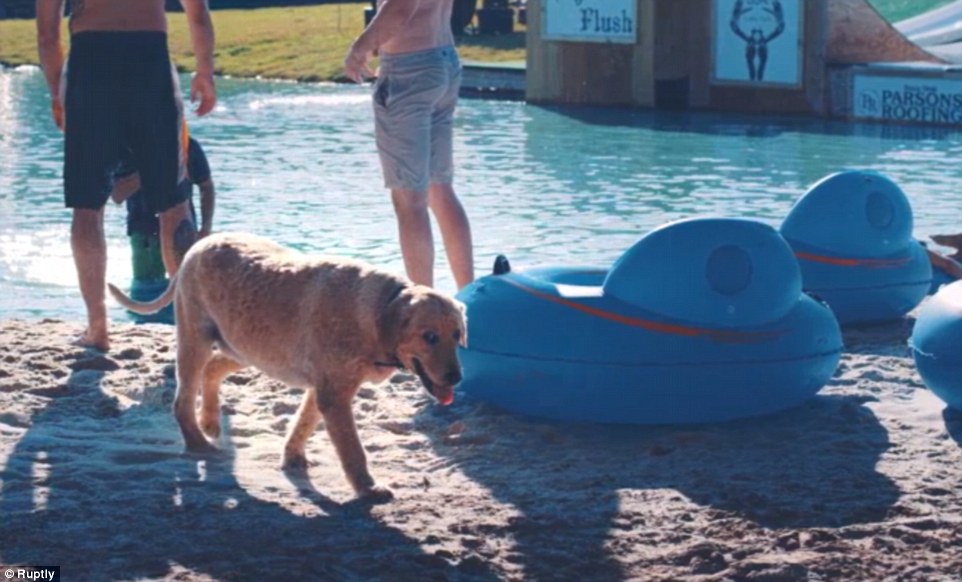}
\caption{coherence-aware \textit{Story}: how to spend a day.\\ coherence--agnostic: dogs playing on the beach.}
\end{subfigure}
\caption{Captions generated by the coherence-aware and coherence-agnostic models. \small{(Photo credits: YesVideo; TinnaPong; Sok Chien Lim; GoPro)}}
\label{fig:generation-label-aware}
\end{figure*}

\paragraph{Model}
We model the output caption using a sequence-generation approach based on Transformer Networks \cite{vaswani2017attention}.
The output is the sequence of sub-tokens comprising the target caption.
The input is obtained by concatenating the following features.

\paragraph{Image Features}
    We obtain a 64 dimensional representation for the image using the Graph-RISE \cite{juan2019graph} feature extractor, which employs a ResNet-101 network to classify images into some 40M classes. 
    We do not fine tune this image encoder model.
    We use the 64-dimensional feature available immediately before the classification layer, and embed into the Transformer encoder embedding space using a trainable dense layer.
    
\paragraph{Detected Objects}
    We obtain object labels for the image using Google Cloud Vision API.\footnote{cloud.google.com/vision}
    We represent each label using pre-trained 512-dimensional vectors trained to predict co-occurring objects on web pages, in a similar fashion as the word2vec model \cite{mikolov2013distributed}. We embed each of these into the Transformer encoder embedding space using a trainable dense layer.
    
\paragraph{Coherence relation label}
This is an input label fed at training time, for which we use the inferred coherence relation for the image--caption pair; at inference time, the label input is used to control the information in the generated caption.
Embeddings for the coherence labels are trainable model parameters.
Additionally, the relationship label serves as the start token for the Transformer decoder (Figure~\ref{fig:model-image-captioning}), i.e., it is made available both for the encoder network and directly for the decoder network.
When training and evaluating a coherence-agnostic model, this label is set to a special symbol, such as \emph{NONE}, essentially running the model without coherence information.
For all models described in this paper, the Transformer network has 6 encoder layers, 6 decoder layers, 8 attention heads, and a 512-dimensional embedding space.

\section{Results and Evaluation}
In what follows, we discuss evidence for our hypotheses: (a) a coherence-aware model presents information that is aligned with the goal of the discourse; and (b) a coherence-aware model can significantly improve caption quality.

\paragraph{Evaluation by expert annotators}
We train the model described above with the predicted discourse relation labels for image--caption pairs in the Conceptual Captions training and validation sets.
The checkpoint with highest CIDEr \cite{vedantam2015cider} score on the validation set is selected for inference and human evaluations. 
We asked our annotators to annotate a subset of randomly selected image--caption pairs generated by this model. 
These evaluation images were selected from the Conceptual Captions evaluation set based on their predicted coherence label using the single-label classifier (Section~\ref{sec:classification}) on the captions generated by the coherence-agnostic model (Section~\ref{sec:generation}).

According to our sensitivity power analysis, with a sample size of 1500 image--text pairs, 300 in each category, we are able to detect effect sizes as small as 0.1650 with a power and significance level of 95\%.
Table~\ref{tab:discourse-aware} shows the result distributions for the coherence-agnostic and coherence-aware model.  Differences greater than 3\% are statistically significant with ($p<0.05, t>2.5$). 
The ability to control the generated caption using an input coherence relation is clear: when asking for \textit{Visible} (the column under Visible), 79.85\% of the captions are evaluated to fit the \textit{Visible} label (non-overlapping), an absolute increase of 27.7\% over the coherence-agnostic model (with only 52.09\% Visible); at the same time, the rate of \textit{Story} and \textit{Subjective} captions reduces significantly.
This reduction is particularly noteworthy in the light of eliminating potential context hallucinations, which are likely to be found under the  \textit{Story} and \textit{Subjective} labels.

A similar trend is observed when asking for, e.g., \textit{Meta}: 46.49\% of the captions are evaluated to fit the \textit{Meta} label (non-overlapping; the column under Meta), up 15.3\% over the coherence-agnostic model (with 31.18\% Story).
A qualitative analysis of the generated captions shows that captions generated under the \textit{Meta} label include terms such as ``screenshot'' and ``view'', while \textit{Subjective} captions come with adjectives such as ``beautiful'' or ``favorite". Figure~\ref{fig:generation-label-aware} shows several examples.

\paragraph{Crowdsouring and Automatic Metrics}
For the following experiments, a subset of the Conceptual Captions validation data was selected where the ground-truth captions are labeled as \textit{Visible}. 


To compare the quality of the generated captions using our framework with other models, we follow the same crowdsourcing protocol that \newcite{sharma2018conceptual} employed for quality assessment.
We asked subjects whether the generated captions are ``good'' or not. 86\% of the captions generated by the coherence-aware model were selected as ``good'' captions, whereas only 74\% of the captions generated by the coherence-agnostic model were selected as ``good'' captions.
Note that, based on the human-evaluation data published\footnote{http://www.conceptualcaptions.com/winners-and-data} for the Conceptual Caption Workshop at CVPR 2019,
this rate is on average 67\% ``good'' captions for the participating state-of-the-art models in 2019. 
Furthermore, in a follow-up experiment we ask subjects to choose between a caption generated by the coherence-aware model and one generated by the coherence-agnostic model:
68.2\% of the time subjects preferred the coherence-aware result, versus 31.8\% for the coherence-agnostic one.

In addition, we study the \textit{quality} and the \textit{relevance} of the captions generated by our model as suggested by \cite{van-der-lee-etal-2019-best}.
On a scale of 0 to 5, the average scores of the \textit{quality} of the captions generated by the coherence-aware and the coherence-agnostic model are, respectively, 3.44 and 2.83.
The average score of the \textit{relevance} for the coherence-aware and the coherence-agnostic conditions are, respectively, 4.43 and 4.40.
Note that subjects rated the quality and the relevance of the captions while seeing the questions on the same page. Screenshots and code for the experiments can be found on our GitHub page. 

With the exception of the \textit{relevance} condition, the results of the other questions that we asked in the crowdsourcing experiments are statistically significantly different ($p < 0.05, t > |3.1|$), which indicates that subjects prefer captions generated by the coherence-aware model.
We also mention here that this difference in quality, albeit significant from a human-rating perspective, is not reflected in the CIDEr score computed on the same data (against the available reference captions).
The CIDEr score of the captions generated by the coherence-aware and the coherence-agnostic models are, respectively, 0.958 and 0.964.
This is not surprising, as the reference captions used by CIDEr are subject to the same distribution over coherence relations as the rest of the data, and therefore generating caption outputs with a different coherence-relation distribution (Table~\ref{tab:discourse-aware}) is unlikely to have a positive impact on reference-driven metrics such as CIDEr.

\section{Conclusions and Future Work}
Representing coherence in image--text presentations can provide a scaffold for organizing, disambiguating and integrating the interpretation of communication across modalities.
We show that cross-modal coherence modeling significantly improves the consistency and quality of the generated text with respect to information needs.
This is a step forward towards designing systems that learn commonsense inferences in images and text and use that to communicate naturally and effectively with the users.
%
In addition, the presented dataset, Clue, provides opportunities for further theoretical and computational explorations.
The experiments described for the coherence relation prediction task set the stage for designing better models for inferring coherence for images--text pairs.

The presented work has limitations that can be addressed in future research.
According to the description of the Conceptual Captions dataset, its captions have been hypernymized. 
However, by studying the examples in the \textit{Other} category, we discovered an additional coherence relation that exists between an image and caption, in which the caption identifies an object or entity in the image--\textit{Identification}.
Examples of this relation involves a caption that mentions the brand of a product or the name of the person in the image.
\textit{Identification} is easy to annotate but missing from this work due to the properties of the corpus we annotated.
Future work should study this additional relation in the context of caption annotation and generation. 

\section{Acknowledgement}

The research presented here is supported by NSF
Awards IIS-1526723 and CCF-19349243 and through a fellowship
from the Rutgers Discovery Informatics Institute.
Thanks to Gabriel Greenberg
and the anonymous reviewers for helpful comments. We would also like to thank the Mechanical Turk annotators for their contributions. We are grateful to our data annotators, Ilana Torres and Kathryn Slusarczyk for their dedicated work.


\bibliographystyle{acl_natbib}

\end{document}